\documentclass[11pt]{article}
\usepackage{acl2016}
\usepackage{times}
\usepackage{latexsym}
\usepackage{graphicx}
\usepackage{color}

\aclfinalcopy 

\title{Character-based Neural Machine Translation}
 
\author{Marta R. Costa-juss\`a \and Jos\'e A. R. Fonollosa \\
         TALP Research Center \\ Universitat Polit\`ecnica de Catalunya, Barcelona \\
  {\tt \{marta.ruiz,jose.fonollosa\}@upc.edu}}

\date{}

\begin{document}

\maketitle

\begin{abstract}
Neural Machine Translation (MT) has reached state-of-the-art results. However, one of the main challenges that neural MT still faces is dealing with very large vocabularies and morphologically rich languages. 

In this paper, we propose a neural MT system using character-based embeddings in combination with convolutional and highway layers to replace the standard lookup-based word representations. The resulting unlimited-vocabulary and affix-aware source word embeddings are tested in a state-of-the-art neural MT based on an attention-based bidirectional recurrent neural network. The proposed MT scheme provides improved results even when the source language is not morphologically rich. 
Improvements up to 3 BLEU points are obtained in the German-English WMT task.
\end{abstract}

\section{Introduction}

Machine Translation (MT) is the set of algorithms that aim at transforming a source language into a target language. For the last 20 years, one of the most popular approaches has been statistical phrase-based MT, which uses a combination of features to maximise the probability of the target sentence given the source sentence \cite{koehn:2003}. Just recently, the neural MT approach has appeared \cite{blunsom:2013,sutskever:2014,cho:2014,Bahdanau:2015} and obtained state-of-the-art results. 

Among its different strengths neural MT does not need to pre-design feature functions beforehand; 
optimizes the entire system at once because it provides a fully trainable model; uses word embeddings \cite{sutskever:2014} so that words (or minimal units) are not independent anymore; and is easily extendable to multimodal sources of information \cite{elliott:2015}. As for weaknesses, neural MT has a strong limitation in vocabulary due to its architecture and it is difficult and computationally expensive to tune all parameters in the deep learning structure. 

In this paper, we use the neural MT baseline system from \cite{Bahdanau:2015}, which follows an encoder-decoder architecture with attention, and introduce elements from the character-based neural language model \cite{kim:2015}. The translation unit continues to be the word, and we continue using word embeddings related to each word as an input vector to the bidirectional recurrent neural network (attention-based mechanism). The difference is that now the embeddings of each word are no longer an independent vector, but are computed from the characters of the corresponding word. The system architecture has changed in that we are using a convolutional neural network (CNN) and a highway network over characters before the attention-based mechanism of the encoder. This is a significant difference from previous work \cite{senrich:2015} which uses the neural MT architecture from \cite{Bahdanau:2015} without modification to deal with sub-word units (but not including unigram characters).

Subword-based representations have already been explored in Natural Language Processing (NLP), e.g. for POS tagging \cite{santos:2014}, name entity recognition \cite{santos:2015}, parsing \cite{ballesteros:2015}, normalization \cite{chrupala:2014} or learning word representations \cite{botha:2014,chen:2015}. These previous works show different advantages of using character-level information. In our case, with the new character-based neural MT architecture, we take advantage of intra-word information, which is proven to be extremely useful in other NLP applications \cite{santos:2014,ling:2015}, especially when dealing with morphologically rich languages. When using the character-based source word embeddings in MT, there ceases to be unknown words in the source input, while the size of the target vocabulary remains unchanged. Although the target vocabulary continues with the same limitation as in the standard neural MT system, the fact that there are no unknown words in the source helps to reduce the number of unknowns in the target. Moreover, the remaining unknown target words can now be more successfully replaced with the corresponding source-aligned words. As a consequence, 
we obtain a significant improvement in terms of translation quality (up to 3 BLEU points). 

The rest of the paper is organized as follows. Section \ref{sec:nmt} briefly explains the architecture of the neural MT that we are using as a baseline system. Section \ref{sec:charnmt} describes the changes introduced in the baseline architecture in order to use character-based embeddings instead of the standard lookup-based word representations. Section \ref{sec:experiments} reports the experimental framework and the results obtained in the German-English WMT task. Finally, section \ref{sec:conclusions} concludes with the contributions of the paper and further work.

\section{Neural Machine Translation}
\label{sec:nmt}

Neural MT uses a neural network approach to compute the conditional probability of the target sentence given the source sentence \cite{cho:2014,Bahdanau:2015}. The approach used in this work \cite{Bahdanau:2015} follows the encoder-decoder architecture.
 First, the encoder reads the source sentence $s=(s_1,..s_I)$ and encodes it into a sequence of hidden states $h=(h_1,..h_I)$. Then, the decoder generates a corresponding translation $t=t_1,...,t_{J}$ based on the encoded sequence of hidden states $h$. Both encoder and decoder are jointly trained to maximize the conditional log-probability of the correct translation.

This baseline autoencoder architecture is improved with a attention-based mechanism \cite{Bahdanau:2015}, in which the encoder uses a bi-directional gated recurrent unit (GRU). This GRU allows for a better performance with long sentences. The decoder also becomes a GRU and each word $t_j$ is predicted based on a recurrent hidden state, the previously  predicted  word $t_{j-1}$, and a context vector. This context vector is obtained from the weighted sum of the annotations $h_k$, which in turn, is computed through an alignment model $\alpha_{jk}$ (a feedforward neural  network). This neural MT approach has achieved competitive results against the standard phrase-based system in the WMT 2015 evaluation \cite{jean:2015}.

%

\section{Character-based Machine Translation}
\label{sec:charnmt}

Word embeddings have been shown to boost the performance in many NLP tasks, including machine translation. However, the standard lookup-based embeddings are limited to a finite-size vocabulary for both computational and sparsity reasons. Moreover, the orthographic representation of the words is completely ignored. The standard learning process is blind to the presence of stems, prefixes, suffixes and any other kind of affixes in words.

As a solution to those drawbacks, new alternative character-based word embeddings have been recently proposed for tasks such as language modeling \cite{kim:2015,ling:2015}, parsing \cite{ballesteros:2015} or POS tagging \cite{ling:2015,santos:2014}. Even in MT \cite{lingb:2015}, where authors use the character transformation presented in \cite{ballesteros:2015,ling:2015} both in the source and target. However, they do not seem to get clear improvements. Recently, \cite{luong:2016} propose a combination of word and characters in neural MT.

\begin{figure}[!ht]
\centering
\includegraphics[width=0.5\textwidth]{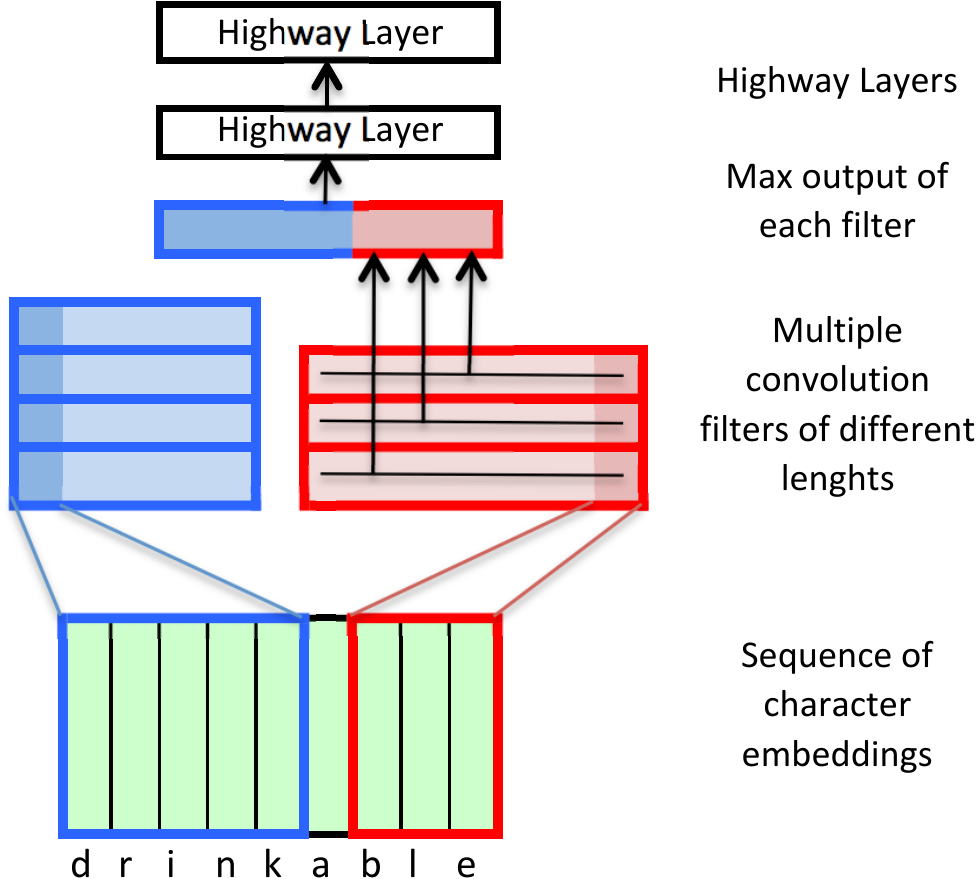}
\caption{Character-based word embedding}
\label{fig:emb}
\end{figure}

For our experiments in neural MT, we selected the best character-based embedding architecture proposed by Kim et al. \cite{kim:2015} for language modeling. As the Figure \ref{fig:emb} shows, the computation of the representation of each word starts with a character-based embedding layer that associates each word (sequence of characters) with a sequence of vectors. This sequence of vectors is then processed with a set of 1D convolution filters of different lengths (from 1 to 7 characters) followed with a max pooling layer. For each convolutional filter, we keep only the output with the maximum value. The concatenation of these max values already provides us with a representation of each word as a vector with a fixed length equal to the total number of convolutional kernels. However, the addition of two highway layers was shown to improve the quality of the language model in \cite{kim:2015} so we also kept these additional layers in our case. The output of the second Highway layer will give us the final vector representation of each source word, replacing the standard source word embedding in the neural machine translation system.

In the target size we are still limited in vocabulary by the softmax layer at the output of the network and we kept the standard target word embeddings in our experiments. However, the results seem to show that the affix-aware representation of the source words has a positive influence on all the components of the network. The global optimization of the integrated model forces the translation model and the internal vector representation of the target words to follow the affix-aware codification of the source words.

\section{Experimental framework}
\label{sec:experiments}

This section reports the data used, its preprocessing, baseline details and results with the enhanced character-based neural MT system.

\subsection{Data}

We used the German-English WMT data\footnote{http://www.statmt.org/wmt15/translation-task.html} including the EPPS, NEWS and Commoncrawl. Pre-processing consisted of tokenizing, truecasing, normalizing punctuation and filtering sentences with more than 5\% of their words in a language other than German or English. Statistics are shown in Table \ref{tab:corpus}. 

\begin{table}[h]
\begin{center} {
\begin{tabular}{|l|l|c|c|c|c|} \hline
L & Set & S & W & V & OOV \\ \hline
De & Train & 3.5M & 77.7M & 1.6M & -\\ 
&Dev & 3k & 63.1k & 13.6k &  1.7k \\
&Test & 2.2k & 44.1k & 9.8k & 1.3k\\ \hline
En & Train & 3.5M & 81.2M & 0.8M & -\\ 
&Dev & 3k & 67.6k & 10.1k  & 0.8k\\
&Test & 2.2k & 46.8k & 7.8k & 0.6k\\ \hline
\end{tabular} }
\end{center}
\caption{Corpus details. Number of sentences (S), words (W), vocabulary (V) and out-of-vocabulary-words (OOV) per set and language (L). M standing for millions, k standing for thousands.}
\label{tab:corpus}
\end{table}

\subsection{Baseline systems}

The phrase-based system was built using Moses \cite{koehn:2007}, with standard parameters such as grow-final-diag for alignment, Good-Turing smoothing of the relative frequencies, 5-gram language modeling using Kneser-Ney discounting, and lexicalized reordering, among others. The neural-based system was built using the software from DL4MT\footnote{http://dl4mt.computing.dcu.ie/} available in github. We generally used settings from previous work \cite{jean:2015}: networks have an embedding of 620 and a dimension of 1024, a batch size of 32, and no dropout.  We used a vocabulary size of 90 thousand words in German-English. Also, as proposed in \cite{jean:2015} we replaced unknown words (UNKs) with the corresponding source word using the alignment information.

\subsection{Results}

\begin{table*}[!ht]
\begin{center} {
\scriptsize
\begin{tabular}{l|l|l}
\hline
1& SRC & Berichten zufolge hofft Indien darüber hinaus auf einen Vertrag zur \textbf{Verteidigungszusammenarbeit} zwischen den beiden Nationen .\\
&Phrase & reportedly hopes India , in addition to a contract for the defence cooperation between the two nations . \\  
&NN & according to reports , India also hopes to establish a contract for the UNK between the two nations .\\  
&CHAR & according to reports , India hopes to see a Treaty of \textbf{Defence Cooperation} between the two nations .\\
&REF & India is also reportedly hoping for a deal on \textbf{defence collaboration} between the two nations .\\
\hline
2&SRC & der durchtrainierte Mainzer sagt von sich , dass er ein `` ambitionierter \textbf{Rennradler} `` ist .\\
&Phrase & the will of Mainz says that he a more ambitious .\\  
&\scriptsize{NN} & the UNK Mainz says that he is a `` ambitious , . ``\\ 
&\scriptsize{CHAR} & the UNK in Mainz says that he is a ' \textbf{ambitious racer} ' . \\ \
&\scriptsize{REF} & the well-conditioned man from Mainz said he was an `` \textbf{ambitious racing cyclist} . ``\\ \hline
3&SRC & die GDL habe jedoch nicht gesagt , wo sie \textbf{streiken} wolle , so dass es schwer sei , die Folgen konkret vorherzusehen .\\
&Phrase & the GDL have , however , not to say , where they strike , so that it is difficult to predict the consequences of concrete .\\  
&NN &however , the UNK did not tell which they wanted to UNK , so it is difficult to predict the consequences .\\
&CHAR & however , the UNK did not say where they wanted to \textbf{strike} , so it is difficult to predict the consequences .\\ 
&REF & the GDL have not said , however , where they will \textbf{strike} , making it difficult to predict exactly what the consequences will be .\\ \hline
4&SRC & die Premierminister Indiens und Japans trafen sich in Tokio .\\
&Phrase & the Prime Minister of India and Japan in Tokyo . \\  
&\scriptsize{NN} & the Prime Minister of India and Japan met in Tokyo \\
&\scriptsize{CHAR} &  the Prime \textbf{Ministers} of India and Japan met in Tokyo \\ 
&\scriptsize{REF} & India and Japan prime \textbf{ministers} meet in Tokyo \\ \hline
5&SRC & wo die Beamten es aus den Augen verloren .\\
&Phrase & where the officials lost sight of\\  
&\scriptsize{NN} & where the officials lost it out of the eyes\\ 
&\scriptsize{CHAR} &  where officials \textbf{lose sight of it}\\ 
&\scriptsize{REF} & causing the officers to \textbf{lose sight of it}\\ \hline
\end{tabular}}
\caption{\label{tab:example}Translation examples.}
\end{center}
\end{table*}

Table \ref{baseline-bleu-table} shows the BLEU results for the baseline systems (including phrase and neural-based, NN) and the character-based neural MT (CHAR). We also include the results for the CHAR and NN systems  with post-processing of unknown words, which consists in replacing the UNKs with the corresponding source word (+Src), as suggested in \cite{jean:2015}. BLEU results improve by almost 1.5 points in German-to-English and by more than 3 points in English-to-German. 
The reduction in the number of unknown words (after postprocessing) goes from 1491 (NN) to 1260 (CHAR) in the direction from German-to-English and from 3148 to 2640 in the opposite direction. 
Note the number of out-of-vocabulary words of the test set is shown in Table \ref{tab:corpus}. 


\begin{table}
\begin{center}
\begin{tabular}{|l|r|r|}
\hline & {De-$>$En} & {En-$>$De} \\ \hline
Phrase & 20.99 & 17.04  \\ \hline
NN & 18.83& 16.47 \\
NN+Src & 20.64 & 17.15 \\ \hline
CHAR & 21.40 & 19.53 \\
CHAR+Src & \textbf{22.10}  & \textbf{20.22} \\ \hline
\end{tabular}
\end{center}
\caption{\label{baseline-bleu-table} De-En BLEU results.}
\end{table}

The character-based embedding has an impact in learning a better translation model at various levels, which seems to include better alignment, reordering, morphological generation and disambiguation. Table \ref{tab:example} shows some examples of the kind of improvements that the character-based neural MT system is capable of achieving compared to baseline systems. Examples 1 and 2 show how the reduction of source unknowns improves the adequacy of the translation. Examples 3 and 4 show how the character-based approach is able to handle morphological variations. Finally, example 5 shows an appropriate semantic disambiguation.



\section{Conclusions}
\label{sec:conclusions}

Neural MT offers a new perspective in the way MT is managed. Its main advantages when compared with previous approaches, e.g. statistical phrase-based, are that the translation is faced with trainable features and optimized in an end-to-end scheme. However, there still remain many challenges left to solve, such as dealing with the limitation in vocabulary size. 

In this paper we have proposed a modification to the standard encoder/decoder neural MT architecture to use unlimited-vocabulary character-based source word embeddings. 
The improvement in BLEU is about 1.5 points in German-to-English and more than 3 points in English-to-German.

As further work, we are currently studying different alternatives \cite{cho:2016} to extend the character-based approach to the target side of the neural MT system.

\section*{Acknowledgements}

This work is supported by the 7th Framework Program of the European Commission through the International Outgoing Fellowship Marie Curie Action (IMTraP-2011-29951) and also by the Spanish Ministerio de Econom\'ia y Competitividad and European Regional Developmend Fund, contract TEC2015-69266-P (MINECO/FEDER, UE).

\bibliography{char-revised}
\bibliographystyle{acl2016}

\end{document}